\documentclass[preprint,12pt]{elsarticle}

\usepackage{amssymb}
\usepackage{amsmath}

\usepackage[table,xcdraw]{xcolor}
\usepackage{colortbl}
\usepackage{amssymb}
\usepackage{multirow} 
\usepackage{makecell}
\usepackage{graphicx}
\usepackage{subcaption}

\usepackage{amssymb} 
\definecolor{warmgreen}{RGB}{0,150,80} 
\definecolor{warmred}{RGB}{220,60,60}    
\newcommand{\uparrowg}{\textcolor{warmgreen}{\large\ensuremath{\blacktriangle}}}
\newcommand{\downarrowr}{\textcolor{warmred}{\large\ensuremath{\blacktriangledown}}}
\usepackage{xcolor}
\usepackage{pifont} 

\usepackage{hyperref}
\usepackage{booktabs}

\journal{Nuclear Physics B}

\begin{document}

\begin{frontmatter}



\title{Are Heterogeneous Graph Neural Networks Truly Effective for Node Classification? A Causal Perspective}


\author[a]{Xiao Yang}
\author[b,c]{Xuejiao Zhao\corref{cor1}}
\author[a]{Zhiqi Shen}

\address[a]{organization={College of Computing and Data Science, Nanyang Technological University}, 
    country={Singapore}}

\address[b]{organization={Joint NTU-UBC Research Centre of Excellence in Active Living for the Elderly (LILY), Nanyang Technological University}, 
    country={Singapore}}

\address[c]{organization={Alibaba-NTU Singapore Joint Research Institute (ANGEL), Nanyang Technological University}, 
    country={Singapore}}

\cortext[cor1]{Corresponding author.}
\begin{abstract}
Graph neural networks (GNNs) have achieved remarkable success in node classification. Building on this progress, heterogeneous graph neural networks (HGNNs) integrate relation types and node and edge semantics to leverage heterogeneous information. Causal analysis for HGNNs is advancing rapidly, aiming to separate genuine causal effects from spurious correlations. However, whether HGNNs are intrinsically effective {for node classification} remains underexamined, and most studies implicitly assume rather than establish this effectiveness.
In this work, we examine HGNNs {for node classification} from two perspectives: model architecture and heterogeneous information. We conduct a systematic reproduction across 21 datasets and 20 baselines, complemented by comprehensive hyperparameter retuning. To further disentangle the source of performance gains, we {develop a causal mediation analysis framework that treats the introduction of heterogeneous relation information as the treatment, candidate structural properties as mediators, and node classification performance as the outcome. This framework first screens candidate mediators according to their treatment-induced changes and their associations with performance improvement, and then decomposes the total effect into mediated and direct effects.} Our results lead to two conclusions. First, model architecture and complexity have no causal effect on {node classification} performance. Second, heterogeneous information exerts a positive causal effect {primarily through increasing homophily and local-global distribution discrepancy}, which makes node classes more distinguishable. The implementation is publicly available at \href{https://github.com/YXNTU/CausalHGNN}{https://github.com/YXNTU/CausalHGNN}.

\end{abstract}



\end{frontmatter}


\section{Introduction}
Node classification is a cornerstone task in graph learning~\cite{wu2020comprehensive, xia2021graph, bei2025correlation, li2020explain}, with broad applications in various domains~\cite{gong2024gc4nc, zhao2021brain, ali2025binary}, such as recommendation systems~\cite{ma2024cross, sharma2024survey, li2018improving}, risk detection~\cite{yang2025generalizable, du2025identifying}, bioinformatics~\cite{zhao2025medrag, zhao2025smart, li2019graph} and drug development~\cite{qiao2025molecular, qiao2025self}. Graph neural networks (GNNs)~\cite{velickovic2017graph, hamilton2017inductive, zhao2021explainable} have emerged as a powerful framework by leveraging graph topology and node features through message passing~\cite{gilmer2017neural, yang2025unified}. However, traditional GNNs are primarily designed for homogeneous graphs, limiting their ability to model the diverse semantics in real-world networks~\cite{shi2016survey, yang2020heterogeneous}. To capture such heterogeneity, heterogeneous graph neural networks (HGNNs)~\cite{zhang2019heterogeneous, wang2022survey} extend GNNs by incorporating heterogeneous information to represent multiple node and edge types.

From a causal perspective, analyzing learning problems has become increasingly mainstream~\cite{yang2024dual, ban2025llm, chen2025causal}. Recent studies have introduced causal perspectives into heterogeneous graph learning to disentangle genuine causal contributions from spurious correlations~\cite{zhu2022link, jing2024causality}.
For instance, Adhikari et al.~\cite{adhikari2025inferring} applied causal analysis to balance heterogeneous peer influence, Zhao et al.~\cite{zhao2024learning} proposed Dual Weighting Regression to capture neighbor interference and mitigate network bias, and Wu et al.~\cite{wu2025estimating} estimated heterogeneous causal effects to enhance HGNN adaptability to downstream tasks.
These efforts demonstrate that causal analysis can provide valuable insights into the mechanisms of HGNNs.
However, all existing approaches implicitly assume that HGNNs are inherently effective rather than verifying this assumption.
If this premise does not hold, causal conclusions built upon it may be unreliable.
Prior empirical evidence~\cite{lv2021we}, in contrast, shows that HGNNs do not consistently outperform simpler GNNs on heterogeneous graph benchmarks.
Resonating with this, recent rigorous evaluations on homogeneous graphs~\cite{luo2024classic} have demonstrated that classic GCNs are strong baselines that often rival complex state-of-the-art architectures when properly tuned.
This tension leads to a fundamental question: are HGNNs genuinely effective {for node classification}, and if so, what underlying factors drive their performance and how can these contributions be rigorously assessed?

To answer this question {in the context of node classification}, we first clarify the role of model architecture in HGNNs.
We reproduce 20 widely used baselines across 21 heterogeneous graph datasets using their official implementations under a unified evaluation protocol, ensuring that comparisons are not confounded by inconsistent training procedures.
In parallel, we perform extensive hyperparameter tuning for a representative and simple HGNN, RGCN~\cite{schlichtkrull2018modeling}, and compare its optimized results with all reproduced baselines.
With model-related effects controlled, we next focus on performance variations arising from heterogeneous information itself.
To isolate this effect, we fix the RGCN backbone and compare different graph constructions, including a homogeneous reference graph and graph variants with heterogeneous relation information introduced.
We then investigate the mechanisms underlying the observed performance changes through causal mediation analysis.
Instead of assuming a single explanatory factor in advance, we construct a pool of candidate structural mediators, including homophily~\cite{ma2021homophily,zhu2020beyond,luan2024graph,zhu2024impact}, local--global distribution discrepancy~\cite{liu2021node,yehudai2021local,loveland2024performance}, degree, density, reachability, diversity, and feature-related structural consistency.
The candidate mediators are first screened according to whether they are affected by the introduction of heterogeneous relation information and whether their treatment-induced changes are associated with node classification improvement.
We then decompose the total effect of heterogeneous information into mediated effects and the remaining direct effect.

Our empirical results lead to two main conclusions.
First, model architecture and complexity have no causal influence on {node classification} performance.
After comprehensive hyperparameter tuning, RGCN~\cite{schlichtkrull2018modeling} achieves better results than most HGNN baselines, indicating that the advantage is not attributable to architectural design itself.
Second, heterogeneous information plays a positive role in node classification under a fixed model architecture.
Compared with the homogeneous reference structure, introducing heterogeneous relation information consistently improves performance, and the mediation analysis shows that this improvement is transmitted primarily through two class-relevant structural factors: homophily and local--global distribution discrepancy.
These factors make local neighborhoods more label-consistent and more distinguishable from the global label distribution, thereby improving node separability.
Together, these findings establish that{, for node classification,} the advantage of HGNNs originates mainly from the structural information carried by heterogeneity rather than from model complexity alone.

In summary, our main contributions are as follows:
\begin{itemize}
\item We present the most comprehensive benchmark to date for heterogeneous graph {node classification}, covering 20 HGNN baselines across 21 datasets under a unified evaluation protocol.
\item {We establish that model architecture and complexity have no causal effect on node classification performance}, showing that performance gains cannot be attributed to architectural design alone.
\item We identify that heterogeneous information has a positive causal effect on {node classification} performance by strengthening class-relevant structural signals, primarily homophily and the difference between local and global label distributions.
\end{itemize}

\section{Related Work}
\subsection{HGNNs}
{Node classification is one of the most widely studied and representative tasks in heterogeneous graph learning, and many existing HGNNs have been evaluated under this setting.}
These models can be roughly divided into three categories. 
The first category treats heterogeneous graphs as relational graphs, directly modeling information within each semantic. 
Representative methods include RGCN~\cite{schlichtkrull2018modeling}, which applies relation-specific transformations on adjacency matrices, HetGNN~\cite{zhang2019heterogeneous}, which samples neighbors across node types, and HGT~\cite{hu2020heterogeneous}, which incorporates type-specific parameters into a heterogeneous attention mechanism. 
The second category relies on meta-paths to capture semantic dependencies. 
HAN~\cite{wang2019heterogeneous} pioneered this approach by attending over manually defined meta-paths, MAGNN~\cite{fu2020magnn} further enriched semantic encoding with multiple meta-path encoders, and GTN~\cite{yun2019graph} automated the construction of meta-paths via graph transformation layers. 
{The third category emphasizes structure encoding and global/contextual modeling rather than standard relation- or meta-path-based local aggregation.}
For example, NSHE~\cite{zhao2020network} samples heterogeneous subgraphs under multiple modes, while HINormer~\cite{mao2023hinormer} and related Transformer-style models~\cite{yang2023simple} introduce global attention and positional encodings to flexibly model heterogeneity. 

\subsection{Causal Effect Estimation on Graphs}
Causal effect analysis originates from the Rubin Causal Model (RCM)~\cite{rubin1974estimating, sekhon2008neyman} and has been increasingly extended to graphs, where both structural dependencies and confounding pose unique challenges. Cotta et al.~\cite{cotta2023causal} introduced causal lifting for link prediction, leveraging invariances to answer counterfactual queries and explain how graph symmetries govern causal link formation. Chen et al.~\cite{chen2023causality} proposed CIE, which applies backdoor adjustment to disentangle causal and spurious features, thereby explaining how data biases in graphs influence node classification. Sun et al.~\cite{sun2024cegrl} further extended this direction to temporal knowledge graphs with CEGRL-TKGR, which disentangles causal and confounding representations via intervention to explain how event relationships drive link prediction. Beyond representation disentanglement, a growing body of work has focused on estimating treatment effects under network interference. Jiang et al.~\cite{jiang2022estimating} estimated individual, peer, and total treatment effects from networked observational data, while Wu et al.~\cite{wu2025estimating} employed orthogonal causal learning to capture heterogeneous direct and spillover effects. Du et al.~\cite{du2024estimating} used causal mediation analysis to disentangle peer direct, peer indirect, and self-treatment effects, and Adhikari et al.~\cite{adhikari2025inferring} inferred direct individual causal effects under heterogeneous peer influence. Several methods explicitly address bias reduction in effect estimation: Farzam et al.~\cite{DBLP:conf/icml/FarzamTS24} proposed geometric causal analysis with Ricci curvature to improve ITE estimation, Cai et al.~\cite{cai2023generalization} derived causal generalization bounds to explain how reweighting mitigates confounding bias, Hu et al.~\cite{hu2025graph} developed disentangled causal analysis to separate adjustment from confounders, Sui et al.~\cite{sui2024invariant} introduced invariant graph learning to capture stable confounders and spillover effects, and Zhao et al.~\cite{zhao2024learning} proposed dual weighting regression to address heterogeneous network interference. Wang~\cite{wang2025large} proposed an LLM-guided divide-and-conquer framework for large-scale causal discovery. Beyond methodological advances, causal effect analysis has also been applied to domain-specific tasks: Ban~\cite{ban2025harnessing} proposed an expert-free causal discovery method leveraging knowledge graph priors. Begum et al.~\cite{begum2025dynamic} designed a Granger causality-inspired GNN to explain temporal dependencies in IoT traffic, Gao et al.~\cite{gao2024introducing} introduced diminutive causal structures and interchange interventions to improve GNN performance, Khaled et al.~\cite{khaled2024graph} extracted causal dependencies via transfer entropy for traffic prediction. Together, these works demonstrate that causal effect estimation on graphs not only advances methodological foundations for treatment effect estimation under interference, but also provides powerful explanatory tools across diverse graph-based applications.

\section{{Preliminaries}} \label{pre}

\subsection{{Potential Outcomes and Causal Mediation Analysis}} \label{CMAsetting}

{
We use the potential outcomes framework to formalize the causal effect of heterogeneous information on node classification performance. 
Consider a set of observational units indexed by \(i \in \{1,\dots,n\}\). 
Each unit receives a binary treatment \(T_i \in \{0,1\}\), where \(T_i=0\) denotes the control condition and \(T_i=1\) denotes the treatment condition. 
In this work, the treatment corresponds to whether heterogeneous relation information is introduced into the input graph. 
Let \(M_i\) denote a mediator that may be affected by the treatment, and let \(Y_i\) denote the final outcome.
}

{
Under causal mediation analysis, the mediator has two potential values:
\begin{equation}
    M_i(0), \quad M_i(1),
\end{equation}
where \(M_i(0)\) is the mediator value under the control condition and \(M_i(1)\) is the mediator value under the treatment condition. 
The outcome is defined as a function of both the treatment and the mediator:
\begin{equation}
    Y_i(t,m),
\end{equation}
where \(Y_i(t,m)\) denotes the potential outcome that would be observed if unit \(i\) received treatment \(T_i=t\) and the mediator were set to \(m\). 
The observed mediator and outcome are therefore
\begin{equation}
    M_i = T_iM_i(1)+(1-T_i)M_i(0),
\end{equation}
\begin{equation}
    Y_i = Y_i\!\left(T_i,M_i(T_i)\right).
\end{equation}
}

{
The total effect measures the overall change in outcome when moving from the control condition to the treatment condition:
\begin{equation}
    \tau
    =
    \mathbb{E}
    \left[
    Y_i\!\left(1,M_i(1)\right)
    -
    Y_i\!\left(0,M_i(0)\right)
    \right].
\end{equation}
Causal mediation analysis decomposes this total effect into an indirect effect transmitted through the mediator and a direct effect not transmitted through that mediator. 
The average causal mediation effect (ACME) under treatment level \(t\) is defined as
\begin{equation}
    \delta(t)
    =
    \mathbb{E}
    \left[
    Y_i\!\left(t,M_i(1)\right)
    -
    Y_i\!\left(t,M_i(0)\right)
    \right],
    \qquad t\in\{0,1\}.
\end{equation}
This quantity captures how much the outcome changes when the mediator changes from its control value to its treatment value while the treatment level is held fixed.
}

{
The average direct effect (ADE) under mediator condition \(t\) is defined as
\begin{equation}
    \zeta(t)
    =
    \mathbb{E}
    \left[
    Y_i\!\left(1,M_i(t)\right)
    -
    Y_i\!\left(0,M_i(t)\right)
    \right],
    \qquad t\in\{0,1\}.
\end{equation}
This quantity captures the remaining effect of the treatment on the outcome when the mediator is held fixed. 
Accordingly, the total effect can be decomposed as
\begin{equation}
    \tau = \delta(1)+\zeta(0) = \delta(0)+\zeta(1).
\end{equation}
In empirical reporting, we use ACME to denote the mediated effect, ADE to denote the remaining direct effect, and compute the proportion mediated as
\begin{equation}
    \mathrm{Proportion\ Mediated}
    =
    \frac{\mathrm{ACME}}{\mathrm{Total\ Effect}}.
\end{equation}
}

{
In the context of this study, the treatment is whether heterogeneous relation information is added to the input graph, the mediators are candidate graph structural factors, and the outcome is node classification performance. 
This formulation allows us to examine not only whether heterogeneous information improves performance, but also through which structural factors this improvement is transmitted.
}

\begin{figure}[htbp]
    \centering
    \includegraphics[width=0.775\linewidth]{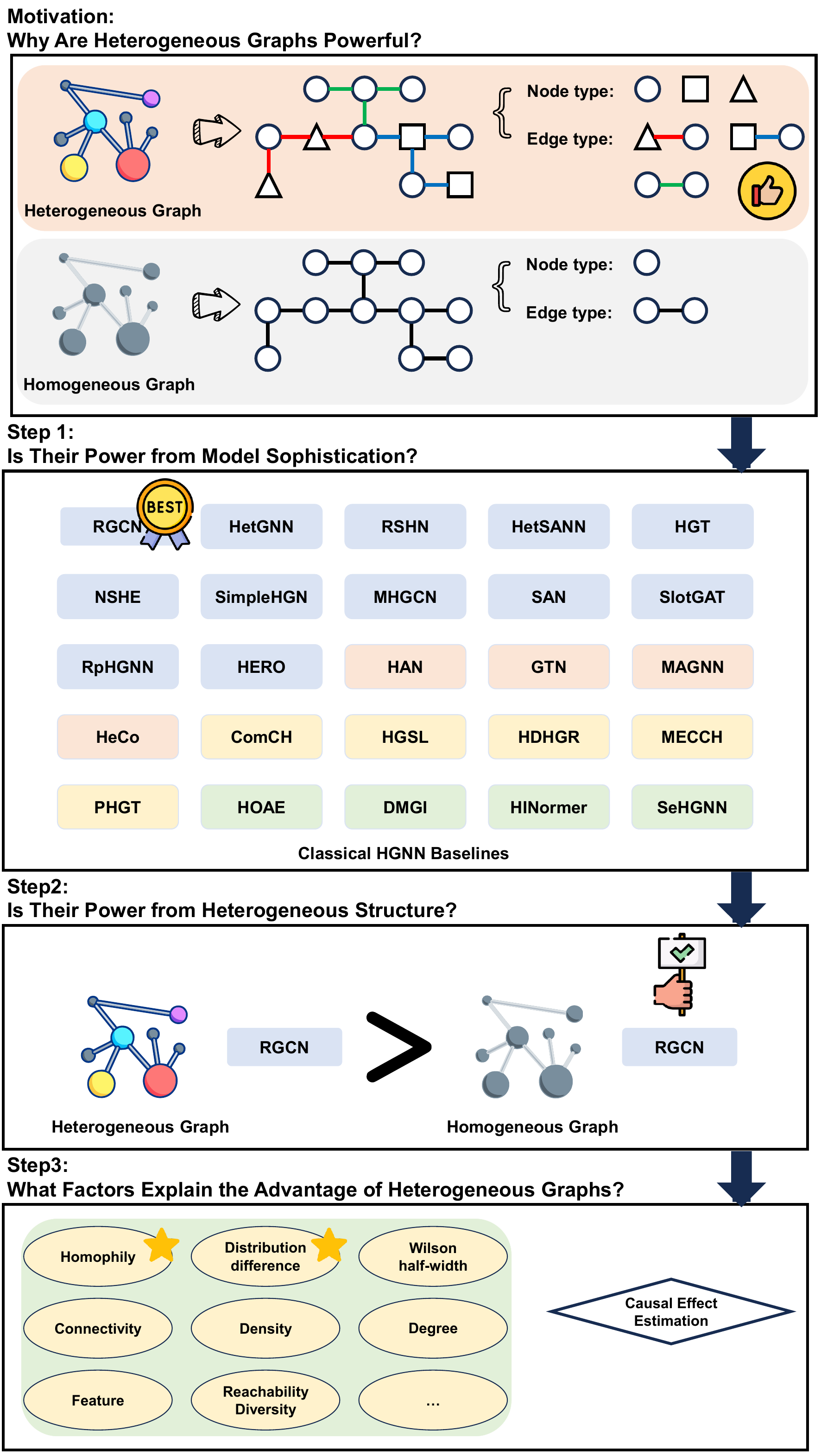}
    \caption{Overview of our research roadmap for disentangling the causal effects of heterogeneous information.}
    \label{fig:overview}
\end{figure}

\section{Establishing and Explaining the Effectiveness of HGNNs {for Node Classification}}
\label{sec:causal-audit}
Figure~\ref{fig:overview} illustrates our three-step causal audit of heterogeneous graph neural networks (HGNNs). 
Step~1 examines whether performance gains arise from architectural sophistication. 
Step~2 tests whether heterogeneous graphs themselves contribute beyond model design. 
Step~3 investigates which structural factors of heterogeneity explain the observed benefits.

\subsection{Step 1: Is performance driven by model architecture?}
Existing studies often take the effectiveness of heterogeneous information for granted and instead focus on designing increasingly sophisticated HGNN architectures. 
This practice risks conflating two potential sources of performance gains: the use of heterogeneous information and the complexity of model design. 
To ensure a valid causal effect estimation, it is necessary to first rule out model architecture as a confounding factor.

We therefore reproduce 20 representative HGNN baselines, summarized in Table~\ref{tab:baseline-overview}, and apply them across 21 heterogeneous graph datasets from nine sources. 
These baselines cover four categories: relation-based, meta-path-based, hybrid (relation \& meta-path), and structure-encoding-based models. 
All implementations strictly follow the official repositories with their reported hyperparameters, while cases containing implementation errors are replaced by the corresponding OpenHGNN implementations~\cite{han2022openhgnn}. 

In addition, motivated by findings in homogeneous graphs where the simplest model (GCN~\cite{kipf2016semi}) can match or even outperform more complex designs~\cite{luo2024classic}, we comprehensively retune the simplest heterogeneous model, RGCN~\cite{schlichtkrull2018modeling}, across all datasets. 
This unified reproduction and hyperparameter retuning enable a fair comparison between RGCN and more sophisticated HGNNs, isolating the causal effect of model architecture from that of heterogeneous information.

\begin{table}[htbp]
\centering
\resizebox{0.9\textwidth}{!}{
\begin{tabular}{ll}
\hline
Type & Model \\
\hline
\multirow{3}{*}{Relation}
& HetGNN~\cite{zhang2019heterogeneous}, RSHN~\cite{zhu2019relation}, SimpleHGN~\cite{lv2021we}, MHGCN~\cite{yu2022multiplex} \\
& SlotGAT~\cite{zhou2023slotgat}, RpHGNN~\cite{hu2024efficient}, HERO$^{\dagger}$~\cite{moself}, SAN~\cite{zhao2022heterogeneous} \\
& HetSANN~\cite{hong2020attention}, HGT~\cite{hu2020heterogeneous}, NSHE~\cite{zhao2020network} \\
\hline
Meta-path
& HAN~\cite{wang2019heterogeneous}, GTN~\cite{yun2019graph}, MAGNN~\cite{fu2020magnn}, HeCo$^{\dagger}$~\cite{wang2021self} \\
\hline
\multirow{2}{*}{Relation \& meta-path}
& ComCH~\cite{li2021leveraging}, HGSL~\cite{zhao2021heterogeneous}, HDHGR~\cite{guo2023homophily} \\
& MECCH~\cite{fu2024mecch}, PHGT~\cite{lu2024heterogeneous} \\
\hline
Structure-encoding-based
& HOAE~\cite{li2024higher}, DMGI$^{\dagger}$~\cite{park2020unsupervised}, HINormer~\cite{mao2023hinormer}, SeHGNN~\cite{yang2023simple} \\
\hline
\end{tabular}}
\caption{
Reproduction of HGNN baselines.
All models are reproduced strictly following their official implementations with paper-reported hyperparameters.
For Relation models, OpenHGNN~\cite{han2022openhgnn} is used when the official repository contains implementation errors.
$^{\dagger}$ denotes self-supervised HGNN models.
}
\label{tab:baseline-overview}
\end{table}

\begin{table}[htbp]
\centering
\resizebox{\textwidth}{!}{%
\begin{tabular}{c|cccccccccc}
\hline
\rowcolor[HTML]{FFF7E6}
& \multicolumn{2}{c}{\cellcolor[HTML]{FFF7E6}HetGNN} & \multicolumn{2}{c}{\cellcolor[HTML]{FFF7E6}RSHN} & SAN & \multicolumn{2}{c}{\cellcolor[HTML]{FFF7E6}HetSANN} & \multicolumn{3}{c}{\cellcolor[HTML]{FFF7E6}GTN} \\ \hline
Source & \multicolumn{2}{c}{Academic} & AIFB & MUTAG & Alibaba & \multicolumn{2}{c}{IMDB} & ACM & IMDB & DBLP \\ \hline
Metric & MaF1 & MiF1 & Acc & Acc & MaF1 & MaF1 & MiF1 & MaF1 & MaF1 & MaF1 \\ \hline
model* & 97.80 & 97.90 & 97.22 & 82.35 & 45.14 & 73.86 & 72.00 & 92.68 & 60.92 & 94.18 \\
RGCN* & - & - & 91.67 & 72.06 & 22.71 & 67.13 & 62.58 & - & - & - \\ \hline
model & 97.80 & 97.90 & 97.22 & 82.35 & 45.10 & 72.51\downarrowr & 70.06\downarrowr & 92.21 & 57.37\downarrowr & 92.93\downarrowr \\
RGCN & \textbf{98.47} & \textbf{98.48} & \textbf{100}\uparrowg & \textbf{84.58}\uparrowg & \textbf{56.30}\uparrowg & \textbf{73.71}\uparrowg & \textbf{71.14}\uparrowg & \textbf{92.63} & \textbf{59.69} & \textbf{93.87} \\ \hline
\rowcolor[HTML]{FFF7E6}
& \multicolumn{2}{c}{\cellcolor[HTML]{FFF7E6}NSHE} & \multicolumn{4}{c}{\cellcolor[HTML]{FFF7E6}SimpleHGN} & \multicolumn{2}{c}{\cellcolor[HTML]{FFF7E6}MHGCN} & \multicolumn{2}{c}{\cellcolor[HTML]{FFF7E6}RpHGNN} \\ \hline
Source & \multicolumn{2}{c}{ACM} & \multicolumn{2}{c}{ACM} & \multicolumn{2}{c}{DBLP} & \multicolumn{2}{c}{DBLP} & \multicolumn{2}{c}{ACM} \\ \hline
Metric & MaF1 & MiF1 & MaF1 & MiF1 & MaF1 & MiF1 & MaF1 & MiF1 & MaF1 & MiF1 \\ \hline
model* & 83.27 & 84.12 & 93.42 & 93.35 & 94.01 & 94.46 & 94.50 & 95.20 & 94.09 & 94.04 \\
RGCN* & - & - & 91.55 & 91.41 & 91.52 & 92.07 & - & - & 91.55 & 91.41 \\ \hline
model & 84.78\uparrowg & 84.95\uparrowg & 93.23 & 93.15 & 93.75 & 94.18 & 93.17\downarrowr & \textbf{93.75}\downarrowr & 93.44 & 93.37 \\
RGCN & \textbf{93.21} & \textbf{93.10} & \textbf{94.11}\uparrowg & \textbf{94.02}\uparrowg & \textbf{94.65}\uparrowg & \textbf{95.00}\uparrowg & \textbf{93.18} & \textbf{93.75} & \textbf{94.11}\uparrowg & \textbf{94.02}\uparrowg \\ \hline
\rowcolor[HTML]{FFF7E6}
& \multicolumn{6}{c}{\cellcolor[HTML]{FFF7E6}HERO} & \multicolumn{2}{c}{\cellcolor[HTML]{FFF7E6}HAN} & \multicolumn{2}{c}{\cellcolor[HTML]{FFF7E6}MAGNN} \\ \hline
Source & \multicolumn{2}{c}{ACM} & \multicolumn{2}{c}{Yelp} & \multicolumn{2}{c}{DBLP} & \multicolumn{2}{c}{ACM} & \multicolumn{2}{c}{DBLP} \\ \hline
Metric & MaF1 & MiF1 & MaF1 & MiF1 & MaF1 & MiF1 & MaF1 & MiF1 & MaF1 & MiF1 \\ \hline
model* & 92.20 & 92.10 & 92.40 & 92.30 & 93.80 & 94.40 & 91.89 & 91.85 & 93.13 & 93.61 \\
RGCN* & - & - & - & - & - & - & - & - & - & - \\ \hline
model & 91.52 & 91.44 & 92.58\uparrowg & 92.20 & 93.11 & 93.97 & 90.63\downarrowr & 90.61\downarrowr & 92.81 & 93.36 \\
RGCN & \textbf{92.55} & \textbf{92.46} & \textbf{93.25} & \textbf{92.84} & \textbf{93.94} & \textbf{94.71} & \textbf{92.45} & \textbf{92.50} & \textbf{93.06} & \textbf{93.59} \\ \hline
\rowcolor[HTML]{FFF7E6}
& \multicolumn{6}{c}{\cellcolor[HTML]{FFF7E6}HGSL} & \multicolumn{4}{c}{\cellcolor[HTML]{FFF7E6}PHGT} \\ \hline
Source & \multicolumn{2}{c}{ACM} & \multicolumn{2}{c}{Yelp} & \multicolumn{2}{c}{DBLP} & \multicolumn{2}{c}{ACM} & \multicolumn{2}{c}{DBLP} \\ \hline
Metric & MaF1 & MiF1 & MaF1 & MiF1 & MaF1 & MiF1 & MaF1 & MiF1 & MaF1 & MiF1 \\ \hline
model* & 93.48 & 93.37 & 93.55 & 92.76 & 91.92 & 92.77 & 93.79 & 93.72 & 94.96 & 95.33 \\
RGCN* & - & - & - & - & - & - & 91.55 & 91.41 & 91.52 & 92.07 \\ \hline
model & \textbf{92.95} & \textbf{92.84} & 92.83 & 92.20 & \textbf{91.68} & \textbf{92.53} & 84.78\downarrowr & 85.62\downarrowr & 93.83\downarrowr & 94.44 \\
RGCN & 92.63 & 92.54 & \textbf{93.25} & \textbf{92.84} & 91.16 & 92.06 & \textbf{94.11}\uparrowg & \textbf{94.02}\uparrowg & \textbf{94.65}\uparrowg & \textbf{95.00}\uparrowg \\ \hline
\rowcolor[HTML]{FFF7E6}
& \multicolumn{2}{c}{\cellcolor[HTML]{FFF7E6}HOAE} & \multicolumn{6}{c}{\cellcolor[HTML]{FFF7E6}DMGI} & \multicolumn{2}{c}{\cellcolor[HTML]{FFF7E6}HINormer} \\ \hline
Source & \multicolumn{2}{c}{ACM} & \multicolumn{2}{c}{ACM} & \multicolumn{2}{c}{IMDB} & \multicolumn{2}{c}{Amazon} & \multicolumn{2}{c}{DBLP} \\ \hline
Metric & MaF1 & MiF1 & MaF1 & MiF1 & MaF1 & MiF1 & MaF1 & MiF1 & MaF1 & MiF1 \\ \hline
model* & 93.11 & 93.03 & 89.80 & 89.80 & 64.80 & 64.80 & 74.60 & 74.80 & 94.57 & 94.94 \\
RGCN* & - & - & - & - & - & - & - & - & 91.52 & 92.07 \\ \hline
model & 93.10 & 93.05 & 84.72\downarrowr & 84.78\downarrowr & 61.18\downarrowr & 61.64\downarrowr & 73.42 & 74.01 & 94.02 & 94.45 \\
RGCN & \textbf{93.21} & \textbf{93.10} & \textbf{89.49} & \textbf{89.39} & \textbf{63.48} & \textbf{63.95} & \textbf{73.76} & \textbf{74.45} & \textbf{94.65}\uparrowg & \textbf{95.00}\uparrowg \\ \hline
\rowcolor[HTML]{FFF7E6}
& \multicolumn{4}{c}{\cellcolor[HTML]{FFF7E6}HGT} & \multicolumn{4}{c}{\cellcolor[HTML]{FFF7E6}SlotGAT} & \multicolumn{2}{c}{\cellcolor[HTML]{FFF7E6}HDHGR} \\ \hline
Source & \multicolumn{2}{c}{IMDB} & \multicolumn{2}{c}{DBLP} & \multicolumn{2}{c}{ACM} & \multicolumn{2}{c}{DBLP} & \multicolumn{2}{c}{IMDB} \\ \hline
Metric & MaF1 & MiF1 & MaF1 & MiF1 & MaF1 & MiF1 & MaF1 & MiF1 & MaF1 & MiF1 \\ \hline
model* & 49.18 & 49.37 & 86.46 & 87.23 & 93.99 & 94.06 & 94.95 & 95.31 & \textbf{58.97} & 59.32 \\
RGCN* & - & - & - & - & 91.55 & 91.41 & 91.52 & 92.07 & 50.33 & 52.51 \\ \hline
model & 49.21 & 49.51 & 86.40 & 87.28 & 93.95 & 94.00 & \textbf{94.86} & \textbf{95.27} & - & - \\
RGCN & \textbf{58.97} & \textbf{59.51} & \textbf{93.06} & \textbf{93.59} & \textbf{94.11}\uparrowg & \textbf{94.02}\uparrowg & 94.65\uparrowg & 95.00\uparrowg & \textbf{58.97}\uparrowg & \textbf{59.51}\uparrowg \\ \hline
\rowcolor[HTML]{FFF7E6}
& \multicolumn{2}{c}{\cellcolor[HTML]{FFF7E6}MECCH} & \multicolumn{2}{c}{\cellcolor[HTML]{FFF7E6}HeCo} & \multicolumn{2}{c}{\cellcolor[HTML]{FFF7E6}ComCH} & \multicolumn{4}{c}{\cellcolor[HTML]{FFF7E6}SeHGNN} \\ \hline
Source & \multicolumn{2}{c}{DBLP} & \multicolumn{2}{c}{ACM} & \multicolumn{2}{c}{DBLP} & \multicolumn{2}{c}{ACM} & \multicolumn{2}{c}{DBLP} \\ \hline
Metric & MaF1 & MiF1 & MaF1 & MiF1 & MaF1 & MiF1 & MaF1 & MiF1 & MaF1 & MiF1 \\ \hline
model* & 94.34 & 95.08 & 89.04 & 88.71 & 94.29 & 94.70 & 94.05 & 93.98 & 95.06 & 95.42 \\
RGCN* & 92.46 & 93.41 & - & - & - & - & 91.55 & 91.41 & 91.52 & 92.07 \\ \hline
model & \textbf{93.99} & \textbf{94.75} & 88.66 & 88.35 & 93.66 & 93.91 & 93.82 & 93.74 & \textbf{94.90} & \textbf{95.26} \\
RGCN & 93.87\uparrowg & 94.66\uparrowg & \textbf{93.21} & \textbf{93.10} & \textbf{94.35} & \textbf{94.87} & \textbf{94.11}\uparrowg & \textbf{94.02}\uparrowg & 94.65\uparrowg & 95.00\uparrowg \\ \hline
\end{tabular}
}
\caption{Reproduction results of 20 HGNN baselines on 21 {node classification} datasets. For each model, {model*} and {RGCN*} denote the results reported in the original papers, {model} denotes our reproduction using the official implementation and reported hyperparameters, and {RGCN} denotes our hyperparameter tuned results. \textbf{Abbreviations:} MaF1 = Macro F1, MiF1 = Micro F1, Acc = Accuracy.}
\label{HGNNbaselinesresults}
\end{table}

\begin{table}[htbp]
\centering
\resizebox{\textwidth}{!}{%
\begin{tabular}{c|cccccccccc}
\hline
\rowcolor[HTML]{FFE4E1}
& \multicolumn{2}{c}{\cellcolor[HTML]{FFE4E1}HetGNN} & \multicolumn{2}{c}{\cellcolor[HTML]{FFE4E1}RSHN} & SAN & \multicolumn{2}{c}{\cellcolor[HTML]{FFE4E1}HetSANN} & \multicolumn{3}{c}{\cellcolor[HTML]{FFE4E1}GTN} \\ \hline
Source & \multicolumn{2}{c}{Academic} & AIFB & MUTAG & Alibaba & \multicolumn{2}{c}{IMDB} & ACM & IMDB & DBLP \\ \hline
Metric & MaF1 & MiF1 & Acc & Acc & MaF1 & MaF1 & MiF1 & MaF1 & MaF1 & MaF1 \\ \hline
model & 97.80 & 97.90 & 97.22 & 82.35 & 45.10 & 72.51 & 70.06 & 92.21 & 57.37 & 92.93 \\
model+   & 98.00 & 98.13 & 97.66  & 82.40 & 48.53  &72.58  &71.10  &92.55  &58.73  &93.51 \\
RGCN & \textbf{98.47} & \textbf{98.48} & \textbf{100} & \textbf{84.58} & \textbf{56.30} & \textbf{73.71} & \textbf{71.14} & \textbf{92.63} & \textbf{59.69} & \textbf{93.87} \\ \hline
\rowcolor[HTML]{FFE4E1}
& \multicolumn{2}{c}{\cellcolor[HTML]{FFE4E1}NSHE} & \multicolumn{4}{c}{\cellcolor[HTML]{FFE4E1}SimpleHGN} & \multicolumn{2}{c}{\cellcolor[HTML]{FFE4E1}MHGCN} & \multicolumn{2}{c}{\cellcolor[HTML]{FFE4E1}RpHGNN} \\ \hline
Source & \multicolumn{2}{c}{ACM} & \multicolumn{2}{c}{ACM} & \multicolumn{2}{c}{DBLP} & \multicolumn{2}{c}{DBLP} & \multicolumn{2}{c}{ACM} \\ \hline
Metric & MaF1 & MiF1 & MaF1 & MiF1 & MaF1 & MiF1 & MaF1 & MiF1 & MaF1 & MiF1 \\ \hline
model & 84.78 & 84.95 & 93.23 & 93.15 & 93.75 & 94.18 & 93.17 & \textbf{93.75} & 93.44 & 93.37 \\
model+  &  85.37& 86.01 &  93.89& 94.00 & 94.05 & 94.56 &  93.17 & \textbf{93.75} & 94.08 & 93.97 \\
RGCN & \textbf{93.21} & \textbf{93.10} & \textbf{94.11} & \textbf{94.02} & \textbf{94.65} & \textbf{95.00} & \textbf{93.18} & \textbf{93.75} & \textbf{94.11} & \textbf{94.02} \\ \hline
\rowcolor[HTML]{FFE4E1}
& \multicolumn{6}{c}{\cellcolor[HTML]{FFE4E1}HERO} & \multicolumn{2}{c}{\cellcolor[HTML]{FFE4E1}HAN} & \multicolumn{2}{c}{\cellcolor[HTML]{FFE4E1}MAGNN} \\ \hline
Source & \multicolumn{2}{c}{ACM} & \multicolumn{2}{c}{Yelp} & \multicolumn{2}{c}{DBLP} & \multicolumn{2}{c}{ACM} & \multicolumn{2}{c}{DBLP} \\ \hline
Metric & MaF1 & MiF1 & MaF1 & MiF1 & MaF1 & MiF1 & MaF1 & MiF1 & MaF1 & MiF1 \\ \hline
model & 91.52 & 91.44 & 92.58 & 92.20 & 93.11 & 93.97 & 90.63 & 90.61 & 92.81 & 93.36 \\
model+  &91.55  &91.56  &92.70  &92.44  & 93.21 &94.05  &92.10  & 92.11 &92.85  & 93.38 \\
RGCN & \textbf{92.55} & \textbf{92.46} & \textbf{93.25} & \textbf{92.84} & \textbf{93.94} & \textbf{94.71} & \textbf{92.45} & \textbf{92.50} & \textbf{93.06} & \textbf{93.59} \\ \hline
\rowcolor[HTML]{FFE4E1}
& \multicolumn{6}{c}{\cellcolor[HTML]{FFE4E1}HGSL} & \multicolumn{4}{c}{\cellcolor[HTML]{FFE4E1}PHGT} \\ \hline
Source & \multicolumn{2}{c}{ACM} & \multicolumn{2}{c}{Yelp} & \multicolumn{2}{c}{DBLP} & \multicolumn{2}{c}{ACM} & \multicolumn{2}{c}{DBLP} \\ \hline
Metric & MaF1 & MiF1 & MaF1 & MiF1 & MaF1 & MiF1 & MaF1 & MiF1 & MaF1 & MiF1 \\ \hline
model & \textbf{92.95} & \textbf{92.84} & 92.83 & 92.20 & \textbf{91.68} & \textbf{92.53} & 84.78 & 85.62 & 93.83 & 94.44 \\
model+  & \textbf{92.95}  & \textbf{92.84} & 92.83  & 92.20 & \textbf{91.68}  & \textbf{92.53} & 85.79 & 87.07 & 93.99 &  94.56\\
RGCN & 92.63 & 92.54 & \textbf{93.25} & \textbf{92.84} & 91.16 & 92.06 & \textbf{94.11} & \textbf{94.02} & \textbf{94.65} & \textbf{95.00} \\ \hline
\rowcolor[HTML]{FFE4E1}
& \multicolumn{2}{c}{\cellcolor[HTML]{FFE4E1}HOAE} & \multicolumn{6}{c}{\cellcolor[HTML]{FFE4E1}DMGI} & \multicolumn{2}{c}{\cellcolor[HTML]{FFE4E1}HINormer} \\ \hline
Source & \multicolumn{2}{c}{ACM} & \multicolumn{2}{c}{ACM} & \multicolumn{2}{c}{IMDB} & \multicolumn{2}{c}{Amazon} & \multicolumn{2}{c}{DBLP} \\ \hline
Metric & MaF1 & MiF1 & MaF1 & MiF1 & MaF1 & MiF1 & MaF1 & MiF1 & MaF1 & MiF1 \\ \hline
model & 93.10 & 93.05 & 84.72 & 84.78 & 61.18 & 61.64 & 73.42 & 74.01 & 94.02 & 94.45 \\
model+  & 93.10 & 93.05  & 85.10  &85.23  & 63.07 & 63.51 & 73.66 & 74.30 &94.57  & 94.89 \\
RGCN & \textbf{93.21} & \textbf{93.10} & \textbf{89.49} & \textbf{89.39} & \textbf{63.48} & \textbf{63.95} & \textbf{73.76} & \textbf{74.45} & \textbf{94.65} & \textbf{95.00} \\ \hline
\rowcolor[HTML]{FFE4E1}
& \multicolumn{4}{c}{\cellcolor[HTML]{FFE4E1}HGT} & \multicolumn{4}{c}{\cellcolor[HTML]{FFE4E1}SlotGAT} & \multicolumn{2}{c}{\cellcolor[HTML]{FFE4E1}HDHGR} \\ \hline
Source & \multicolumn{2}{c}{IMDB} & \multicolumn{2}{c}{DBLP} & \multicolumn{2}{c}{ACM} & \multicolumn{2}{c}{DBLP} & \multicolumn{2}{c}{IMDB} \\ \hline
Metric & MaF1 & MiF1 & MaF1 & MiF1 & MaF1 & MiF1 & MaF1 & MiF1 & MaF1 & MiF1 \\ \hline
model & 49.21 & 49.51 & 86.40 & 87.28 & 93.95 & 94.00 & {94.86} & {95.27} & - & - \\
model+  &  53.45&53.66  &89.97  &90.30  &94.10  & \textbf{94.11}  & \textbf{94.99} & \textbf{95.48} & - & - \\
RGCN & \textbf{58.97} & \textbf{59.51} & \textbf{93.06} & \textbf{93.59} & \textbf{94.11} & 94.02 & 94.65 & 95.00 & \textbf{58.97} & \textbf{59.51} \\ \hline
\rowcolor[HTML]{FFE4E1}
& \multicolumn{2}{c}{\cellcolor[HTML]{FFE4E1}MECCH} & \multicolumn{2}{c}{\cellcolor[HTML]{FFE4E1}HeCo} & \multicolumn{2}{c}{\cellcolor[HTML]{FFE4E1}ComCH} & \multicolumn{4}{c}{\cellcolor[HTML]{FFE4E1}SeHGNN} \\ \hline
Source & \multicolumn{2}{c}{DBLP} & \multicolumn{2}{c}{ACM} & \multicolumn{2}{c}{DBLP} & \multicolumn{2}{c}{ACM} & \multicolumn{2}{c}{DBLP} \\ \hline
Metric & MaF1 & MiF1 & MaF1 & MiF1 & MaF1 & MiF1 & MaF1 & MiF1 & MaF1 & MiF1 \\ \hline
model & {93.99} & {94.75} & 88.66 & 88.35 & 93.66 & 93.91 & 93.82 & 93.74 & {94.90} & {95.26} \\
model+  & \textbf{94.45}  & \textbf{95.10} & 89.03 &88.66  &93.91  & 94.28  &94.10  & 94.00  & \textbf{95.11}  &  \textbf{95.38} \\
RGCN & 93.87 & 94.66 & \textbf{93.21} & \textbf{93.10} & \textbf{94.35} & \textbf{94.87} & \textbf{94.11} & \textbf{94.02} & 94.65 & 95.00 \\ \hline
\end{tabular}
}
\caption{Performance comparison of 20 HGNN baselines on 21 {node classification} datasets. For each model, model denotes our reproduction using the official implementation and reported hyperparameters, model\textsuperscript{+} denotes the results when baselines are fully tuned to the same extent as RGCN to ensure fair comparison, and RGCN denotes our hyperparameter tuned results. Abbreviations: MaF1 = Macro F1, MiF1 = Micro F1, Acc = Accuracy.}
\label{HGNNbaselinesnew}
\end{table}

\begin{table}[htbp]
\centering
\resizebox{0.8\textwidth}{!}{
\begin{tabular}{cccccc}
\hline
Source                   & Dataset                   & Metric   & \textit{Homo}   & \textit{Homo+Hete} & \textit{Hete}  \\ \hline
\multirow{9}{*}{ACM}      & \multirow{2}{*}{ACM-A}    & Macro F1 & 85.46 & \underline{87.31} & \textbf{89.49} \\
                          &                           & Micro F1 & 85.74 & \underline{87.18} & \textbf{89.39} \\
                          & ACM-B                     & Macro F1 & 83.58 & \underline{88.74} & \textbf{92.63} \\
                          & \multirow{2}{*}{ACM-C}    & Macro F1 & 82.30 & \underline{88.65} & \textbf{93.21} \\
                          &                           & Micro F1 & 82.44 & \underline{88.41} & \textbf{93.10} \\
                          & \multirow{2}{*}{ACM-D}    & Macro F1 & 89.92 & \underline{91.10} & \textbf{92.55} \\
                          &                           & Micro F1 & 89.94 & \underline{91.03} & \textbf{92.46} \\
                          & \multirow{2}{*}{ACM-E}    & Macro F1 & 91.96 & \underline{93.07} & \textbf{94.11} \\
                          &                           & Micro F1 & 91.88 & \underline{92.95} & \textbf{94.02} \\ \hline

\multirow{5}{*}{IMDB}     & \multirow{2}{*}{IMDB-A}   & Macro F1 & 61.84 & \underline{62.66} & \textbf{63.48} \\
                          &                           & Micro F1 & 62.72 & \underline{63.21} & \textbf{63.95} \\
                          & IMDB-B                    & Macro F1 & 49.38 & \underline{55.12} & \textbf{59.69} \\
                          & \multirow{2}{*}{IMDB-C}   & Macro F1 & 66.97 & \underline{70.28} & \textbf{73.71} \\
                          &                           & Micro F1 & 58.33 & \underline{65.42} & \textbf{71.14} \\ \hline

\multirow{13}{*}{DBLP}    & DBLP-A                    & Macro F1 & 85.88 & \underline{90.41} & \textbf{93.87} \\
                          & \multirow{2}{*}{DBLP-B}   & Macro F1 & 85.82 & \underline{88.73} & \textbf{91.16} \\
                          &                           & Micro F1 & 86.81 & \underline{89.64} & \textbf{92.06} \\
                          & \multirow{2}{*}{DBLP-C}   & Macro F1 & 87.85 & \underline{90.32} & \textbf{93.06} \\
                          &                           & Micro F1 & 88.81 & \underline{91.11} & \textbf{93.59} \\
                          & \multirow{2}{*}{DBLP-D}   & Macro F1 & 88.99 & \underline{91.07} & \textbf{93.18} \\
                          &                           & Micro F1 & 90.10 & \underline{92.01} & \textbf{93.75} \\
                          & \multirow{2}{*}{DBLP-E}   & Macro F1 & 88.52 & \underline{91.53} & \textbf{93.94} \\
                          &                           & Micro F1 & 89.61 & \underline{92.60} & \textbf{94.71} \\
                          & \multirow{2}{*}{DBLP-F}   & Macro F1 & 87.00 & \underline{91.04} & \textbf{94.65} \\
                          &                           & Micro F1 & 88.33 & \underline{92.11} & \textbf{95.00} \\
                          & \multirow{2}{*}{DBLP-G}   & Macro F1 & 92.26 & \underline{93.26} & \textbf{94.35} \\
                          &                           & Micro F1 & 92.31 & \underline{93.56} & \textbf{94.87} \\ \hline

\multirow{2}{*}{Academic} & \multirow{2}{*}{MC}       & Macro F1 & 97.75 & \underline{98.12} & \textbf{98.47} \\
                          &                           & Micro F1 & 97.79 & \underline{98.16} & \textbf{98.48} \\ \hline

\multirow{2}{*}{Yelp}     & \multirow{2}{*}{Yelp}     & Macro F1 & 88.52 & \underline{91.25} & \textbf{93.94} \\
                          &                           & Micro F1 & 89.61 & \underline{92.33} & \textbf{94.71} \\ \hline

Alibaba                   & Event                     & Macro F1 & 42.32 & \underline{49.81} & \textbf{56.30} \\
AIFB                      & AIFB                      & Accuracy & 97.22 & \underline{98.61} & \textbf{100.00} \\
MUTAG                     & MUTAG                     & Accuracy & 79.41 & \underline{82.07} & \textbf{84.58} \\ \hline

\multirow{2}{*}{Amazon}   & \multirow{2}{*}{Amazon-A} & Macro F1 & 68.02 & \underline{70.91} & \textbf{73.76} \\
                          &                           & Micro F1 & 69.16 & \underline{71.82} & \textbf{74.45} \\ \hline
\end{tabular}}
\caption{Comparison of RGCN under different graph structures. \textit{Homo} denotes the homogeneous projection obtained by removing node and edge types from the raw heterogeneous graph. \textit{Homo+Hete} denotes the homogeneous projection augmented with additional heterogeneous relation channels while maintaining the same RGCN backbone. \textit{Hete} denotes the raw heterogeneous graph. The best performance is marked in bold, and the second-best performance is underlined.}
\label{tab:capacity_enhanced_comparison}
\end{table}

The results in Table~\ref{HGNNbaselinesresults} reveal three main findings. 
First, the performance of RGCN has been consistently underestimated. 
With systematic hyperparameter tuning, RGCN achieves substantial improvements across datasets. 
For example, on AIFB the reported accuracy is 91.67, while our tuned RGCN reaches {100} (+8.33); on MUTAG, the reported score is 72.06, compared to {84.58} (+12.52) after tuning. 
These examples illustrate that tuning yields consistent gains beyond previously reported numbers. 
Second, apart from a few marginal cases, the fine-tuned RGCN outperforms the majority of HGNN models. 
For instance, on DBLP the Macro F1 and Micro F1 of SlotGAT (94.86 and 95.27) are slightly higher than those of RGCN (94.65 and 95.00). 
Beyond such small differences, RGCN clearly surpasses most baselines: on DBLP it improves upon HGT by more than six points in both metrics, and on ACM it achieves around nine points higher than PHGT. 
To ensure a strictly fair comparison, we further extended our experiments by applying the same hyperparameter tuning strategies to the baseline models. As shown in Table \ref{HGNNbaselinesnew}, although most baselines achieved significant performance improvements after tuning, none were able to consistently outperform the RGCN.
These results show that once properly tuned, RGCN not only matches but often exceeds more sophisticated HGNN architectures. 
Third, some HGNN models show a clear performance drop when reproduced, yet the tuned RGCN still surpasses even their originally reported results. 
For example, on ACM the reproduced PHGT achieves only 84.78 Macro F1 and 85.62 Micro F1, far below its reported 93.79 and 93.72. 
Nevertheless, our tuned RGCN attains {94.11} Macro F1 and {94.02} Micro F1, higher than the original PHGT numbers, reinforcing that model architecture is not the decisive factor. 

In summary, the superior performance observed on heterogeneous graphs cannot be attributed to architectural complexity. 
After rigorous hyperparameter optimization, RGCN as the simplest HGNN architecture consistently outperforms most competing models and in many cases even surpasses the published results of advanced HGNNs. 
This motivates us to shift focus from model design to the causal contribution of heterogeneous information itself.

\subsection{{Step 2: Does heterogeneous information improve node classification?}}
{
Having established in Step~1 that model architecture is not the primary source of performance gains, we now turn to the role of heterogeneous information itself. A key requirement for isolating this effect is to avoid changing the model architecture while changing the graph structure. Therefore, instead of comparing different model families, we adopt a controlled experimental design in which the RGCN backbone is fixed across all settings, and only the input graph structure is modified. This allows us to directly evaluate the contribution of heterogeneous information while eliminating architecture as a potential confounding factor.
}

{
For each heterogeneous graph, we construct three graph variants. The first variant, \textit{Homo}, is obtained by transforming the raw heterogeneous graph into its homogeneous projection, where all node and edge types are removed and all relations are merged into a single relation. We do not manually compress or alter raw node features when constructing Homo; only node and edge type information is removed. The second variant, \textit{Homo+Hete}, augments the homogeneous graph with heterogeneous relation information, allowing both homogeneous and heterogeneous structures to be utilized simultaneously. The third variant, \textit{Hete}, corresponds to the original heterogeneous graph, where all relation-specific semantics are preserved. Importantly, all three graph variants are trained using the same RGCN backbone, identical training protocols, and the same hyperparameter search strategy. Consequently, any observed performance differences can be attributed to the graph structure itself rather than to changes in model architecture.
}

{
The results are summarized in Table~\ref{tab:capacity_enhanced_comparison}. Across all datasets and evaluation metrics, a highly consistent trend can be observed. Introducing heterogeneous information substantially improves node classification performance. In nearly all cases, \textit{Homo+Hete} consistently outperforms \textit{Homo}, while \textit{Hete} further achieves the best performance. For example, on ACM-C, the Macro F1 score increases from 82.30 under \textit{Homo} to 88.65 under \textit{Homo+Hete}, and further to 93.21 under \textit{Hete}. On IMDB-C, the Micro F1 score improves from 58.33 to 65.42 and then to 71.14. Similar improvements can be observed across DBLP, Yelp, Amazon, AIFB, MUTAG, and other datasets. These results provide consistent evidence that heterogeneous information is beneficial for node classification.
}

{
An additional observation is that the performance gain cannot be explained solely by introducing more connections or additional message-passing channels. Although \textit{Homo+Hete} consistently improves upon \textit{Homo}, it remains inferior to \textit{Hete} across almost all datasets. If the benefit originated merely from increased connectivity, \textit{Homo+Hete} would be expected to achieve comparable performance to \textit{Hete}. Instead, the results suggest that relation-specific heterogeneous semantics provide structural signals that are not captured by homogeneous aggregation alone. In other words, heterogeneous information appears to reshape the graph in a manner that makes node classes more distinguishable and therefore easier to classify.
}

{
Overall, the results of Step~2 establish that heterogeneous information consistently improves node classification performance under a fixed model architecture. However, this analysis only demonstrates the existence of the effect and does not explain the underlying mechanism. The remaining question is therefore why heterogeneous information is beneficial. In the next section, we investigate how heterogeneous information alters graph structure and identify the structural factors that ultimately contribute to improved node classification performance.
}

\begin{table}[t]
\centering
\small
\begin{tabular}{p{0.28\linewidth}p{0.62\linewidth}}
\hline
\textbf{Hyperparameter} & \textbf{Search space} \\
\hline
Hidden dimension & $\{32, 64, 128, 192, 256, 512, 1024, 4096\}$ \\
Number of layers & $\{0, 1, 2, 3, 4, 5, 6, 7, 8\}$ \\
Residual connection & $\{\mathrm{False}, \mathrm{True}\}$ \\
Feature dropout & $\{0.0, 0.1, 0.2, 0.3, 0.4, 0.5, 0.6\}$ \\
Edge dropout & $\{0.0, 0.05, 0.1, 0.2, 0.3, 0.4, 0.5, 0.6, 0.7, 0.9999\}$ \\
Virtual class dimension & $\{0, 1, 2, 3, 4, 5, 8, 10, 16, 50, 100\}$ \\
Learning rate & $\{10^{-4}, 5 \times 10^{-4}, 10^{-3}, 3 \times 10^{-3}, 5 \times 10^{-3}, 10^{-2}\}$ \\
Weight decay & $\{0, 10^{-6}, 10^{-5}, 10^{-4}, 5 \times 10^{-4}, 10^{-3}, 5 \times 10^{-3}\}$ \\
\hline
\end{tabular}
\caption{Hyperparameter search space used for the tuned baseline setting.}
\label{tab:hyperparameter_search_space}
\end{table}

{Note. To ensure a fair comparison, the {model+} results are obtained by applying the same hyperparameter tuning protocol to all baselines as used for RGCN. Specifically, we perform an exhaustive grid search over the hyperparameter space defined in Table~\ref{tab:hyperparameter_search_space} for all models, whenever the corresponding parameter is supported by their architecture. For model-specific parameters not listed in the table, we strictly follow the official implementations or the settings reported in their original papers. The best configuration for each model is selected based on validation performance and subsequently evaluated on the test set using identical data splits, early stopping criteria, and repeated-run protocols. This ensures that all competing baselines are tuned and optimized to the exact same extent as RGCN.}

\subsection{{Step 3: What factors make heterogeneous information useful for node classification?}}

{
Step~2 shows that heterogeneous information consistently improves node classification performance under a fixed RGCN backbone. 
However, this result only establishes the existence of a performance benefit and does not explain the underlying mechanism. 
In this section, we investigate which structural factors transmit the effect of heterogeneous information to the final prediction outcome. 
We formulate this problem as a causal mediation analysis, where heterogeneous information is treated as an intervention on the input graph, structural properties are considered as candidate mediators, and node-level prediction performance is used as the outcome.
}

{
The central question is whether heterogeneous information improves node classification by changing graph structural properties that make node classes more distinguishable. 
Accordingly, we consider the following mediation pathway:
\begin{equation}
\text{Heterogeneous Information}
\rightarrow
\text{Structural Mediators}
\rightarrow
\text{Prediction Outcome}.
\end{equation}
This formulation separates the existence of heterogeneous information from the prediction result itself. 
The treatment is determined by the graph construction before model training, while the outcome is measured only after the model has been trained and evaluated.
}

\subsubsection{{Treatment, mediators, and outcome}}

{
Formally, the treatment variable is defined as
\begin{equation}
T =
\begin{cases}
0, & \text{RGCN trained on } G_{\mathrm{Homo}}, \\
1, & \text{RGCN trained on } G_{\mathrm{Homo+Hete}}.
\end{cases}
\end{equation}
Here, \(T=0\) denotes the absence of heterogeneous relation information, and \(T=1\) denotes the presence of additional heterogeneous relation information. 
Since \(T\) is determined by the graph construction, it is independent of the realized prediction result.
}

{
We define mediators as structural factors that may be altered by heterogeneous information and may further affect node classification. 
Instead of assuming a specific mediator in advance, we start from a set of candidate structural mediators. 
These candidates include homophily, local--global distribution discrepancy, degree, density, connectivity, reachability, diversity, and other node-level structural indicators. 
For a candidate mediator \(M_k\), its potential values under the two graph conditions are defined as
\begin{equation}
M_k(0,v)=M_k(G_{\mathrm{Homo}},v),
\end{equation}
\begin{equation}
M_k(1,v)=M_k(G_{\mathrm{Homo+Hete}},v).
\end{equation}
The treatment-induced change of this candidate mediator is then computed by
\begin{equation}
\Delta M_k(v)=M_k(1,v)-M_k(0,v).
\end{equation}
A structural factor can serve as a valid mediator only if it is affected by the introduction of heterogeneous information and is also associated with the resulting change in prediction outcome.
}

{
The outcome variable is defined as node classification performance under each graph condition. 
For a graph condition \(T=t\), we train RGCN with 10 random seeds from 0 to 9 and report the average node classification performance across these runs. 
Formally, the outcome under condition \(t\) is defined as
\begin{equation}
Y(t)=\frac{1}{10}\sum_{r=0}^{9}
\mathrm{Perf}\!\left(G_t;\theta^{(r)}\right),
\qquad t\in\{0,1\},
\end{equation}
where \(G_0=G_{\mathrm{Homo}}\), \(G_1=G_{\mathrm{Homo+Hete}}\), \(\theta^{(r)}\) denotes the model trained with random seed \(r\), and \(\mathrm{Perf}(\cdot)\) denotes the node classification metric used for the corresponding dataset. 
Thus, \(Y(0)\) represents the node classification performance when heterogeneous relation information is absent, while \(Y(1)\) represents the performance after heterogeneous relation information is introduced.
}

{
The performance change caused by adding heterogeneous information is measured as
\begin{equation}
\Delta Y = Y(1)-Y(0).
\end{equation}
A positive \(\Delta Y\) indicates that introducing heterogeneous relation information improves node classification performance. 
Under this formulation, Step~3 aims to identify which structural factors explain the performance improvement from \(G_{\mathrm{Homo}}\) to \(G_{\mathrm{Homo+Hete}}\). 
The subsequent analysis first screens candidate mediators according to whether they are affected by heterogeneous information and whether they are associated with node classification performance, and then decomposes the effect of heterogeneous information into indirect effects transmitted through the selected mediators and the remaining direct effect.
}

\subsubsection{{Candidate structural mediators}}

{
Heterogeneous information may influence node classification through multiple structural aspects of the graph. 
Therefore, we do not pre-assume a single mediator. 
Instead, we construct a candidate mediator pool that covers both label-aware structural signals and topology-aware structural signals. 
These candidates are computed under both \(G_{\mathrm{Homo}}\) and \(G_{\mathrm{Homo+Hete}}\), and their treatment-induced changes are later evaluated using a unified mediation screening procedure.
}

{
The candidate mediator pool is defined as
\begin{equation}
\mathcal{M}
=
\{
H,\,
D,\,
W,\,
K,\,
\rho,\,
C,\,
R,\,
S,\,
F
\},
\end{equation}
where each element represents a possible structural pathway through which heterogeneous information may affect node classification performance.
}

{
\textbf{Homophily \(H\).}
Homophily measures the extent to which a node is connected to neighbors with the same class label. 
For node classification, higher homophily usually indicates that neighborhood aggregation provides more class-consistent information. 
Thus, if heterogeneous information increases homophily, it may improve classification by making local neighborhoods more label-consistent.
}

{
\textbf{Local--global distribution discrepancy \(D\).}
This factor measures the difference between a node's local neighborhood label distribution and the global label distribution of the graph. 
A larger discrepancy indicates that the local neighborhood contains more class-specific information rather than merely reflecting the global class prior. 
Therefore, this factor captures whether heterogeneous information makes local structures more discriminative for node classification.
}

{
\textbf{Wilson half-width \(W\).}
Wilson half-width measures the uncertainty of a local label proportion estimate. 
When a node has only a few neighbors or an unstable local label distribution, the corresponding local structural signal is less reliable. 
A smaller Wilson half-width indicates a more reliable local estimate, while a larger value indicates higher uncertainty. 
This factor is included to examine whether heterogeneous information improves node classification by making neighborhood-level label signals more statistically reliable.
}

{
\textbf{Degree \(K\).}
Degree measures the number of neighbors available for message passing. 
Heterogeneous information may change the effective neighborhood size by adding relation-specific channels. 
This factor tests whether the performance gain mainly comes from exposing each node to more neighboring information.
}

{
\textbf{Density \(\rho\).}
Density characterizes how tightly connected a node's local neighborhood is. 
A denser neighborhood may provide stronger local structural context, while a sparse neighborhood may contain weaker aggregation signals. 
This factor is included to test whether heterogeneous information improves classification by increasing local structural compactness.
}

{
\textbf{Connectivity \(C\).}
Connectivity measures whether heterogeneous information improves the accessibility between nodes in the graph. 
If relation-specific edges connect previously weakly connected regions, they may facilitate message passing across useful parts of the graph. 
This factor evaluates whether the benefit of heterogeneous information is mediated by improved graph connectivity.
}

{
\textbf{Reachability \(R\).}
Reachability measures the extent to which nodes can access other nodes within a limited number of hops. 
Compared with degree, reachability captures broader neighborhood coverage rather than immediate local connectivity. 
This factor tests whether heterogeneous information improves classification by expanding the effective receptive field of target nodes.
}

{
\textbf{Diversity \(S\).}
Diversity measures the heterogeneity of the neighborhood composition, such as the variety of relation types, node types, or label categories around a node. 
A more diverse neighborhood may provide richer semantic evidence, but it may also introduce noisy or conflicting information. 
This factor is included to determine whether semantic diversity itself serves as a mediator of the performance gain.
}

{
\textbf{Feature-related structural consistency \(F\).}
Since the raw node features are kept unchanged between \(G_{\mathrm{Homo}}\) and \(G_{\mathrm{Homo+Hete}}\), node features themselves cannot serve as treatment-induced mediators. 
However, heterogeneous information may change how features are aggregated through different neighborhoods. 
We therefore consider feature-related structural consistency, which measures whether a node is connected to neighbors with similar or discriminative feature representations. 
This factor tests whether heterogeneous information improves classification by altering the feature context exposed during message passing.
}

\subsubsection{{Mediator screening}}

{
After constructing the candidate mediator pool, we screen these structural factors before performing causal mediation decomposition. 
A valid mediator should satisfy two necessary conditions. 
First, it should be affected by the treatment, namely the introduction of heterogeneous relation information from \(G_{\mathrm{Homo}}\) to \(G_{\mathrm{Homo+Hete}}\). 
Second, its change should be associated with the improvement in node classification performance. 
Therefore, we conduct a two-step screening procedure. 
The first step examines the treatment-to-mediator relationship \(T \rightarrow M\), and the second step examines the mediator-to-outcome relationship \(M \rightarrow Y\).
}

\begin{table}[htbp]
\centering
{
\resizebox{\textwidth}{!}{
\begin{tabular}{lccc}
\hline
Candidate mediator \(M_k\) 
& Average under \(G_{\mathrm{Homo}}\) 
& Average under \(G_{\mathrm{Homo+Hete}}\) 
& Mean relative change (\%) \\ 
\hline

Global label distribution 
& 0.142 
& 0.142 
& 0.00 \\

Feature consistency \(F\) 
& 0.785 
& 0.788 
& \(+0.38\) \\

Graph density \(\rho\) 
& 0.012 
& 0.015 
& \(+25.00\) \\

Average degree \(K\) 
& 5.34 
& 6.82 
& \(+27.71\) \\

Reachability \(R\) 
& 14.2 
& 16.5 
& \(+16.19\) \\

Neighborhood diversity \(S\) 
& 1.15 
& 2.05 
& \(+78.26\) \\

\textbf{Homophily \(H\)} 
& \textbf{0.421} 
& \textbf{0.518} 
& \textbf{\(+23.04\)} \\

\textbf{Local--global discrepancy \(D\)} 
& \textbf{0.185} 
& \textbf{0.233} 
& \textbf{\(+25.94\)} \\ 
\hline
\end{tabular}
}
}
\caption{{First-step mediator screening based on the treatment-to-mediator relationship \(T \rightarrow M\). 
The table reports the average value of each candidate mediator under \(G_{\mathrm{Homo}}\) and \(G_{\mathrm{Homo+Hete}}\), together with the mean relative change after heterogeneous relation information is introduced.}}
\label{tab:mediator_screening_step1}
\end{table}

{
Table~\ref{tab:mediator_screening_step1} reports the first-step screening results. 
The global label distribution remains unchanged because the node labels and label proportions are identical under \(G_{\mathrm{Homo}}\) and \(G_{\mathrm{Homo+Hete}}\). 
Therefore, it cannot serve as a treatment-induced mediator. 
Feature consistency changes only marginally, with a mean relative increase of \(0.38\%\), indicating that the feature-related context is almost unaffected by adding heterogeneous relation information. 
Thus, these two factors are removed after the first screening step.
}

{
In contrast, several structural factors are clearly affected by heterogeneous information. 
Graph density, average degree, reachability, and neighborhood diversity increase after heterogeneous relation channels are introduced, suggesting that heterogeneous information can change the accessible message-passing structure. 
More importantly, the two label-aware structural factors, homophily and local--global discrepancy, also increase substantially. 
Homophily increases from 0.421 to 0.518, with a mean relative change of \(23.04\%\), while local--global discrepancy increases from 0.185 to 0.233, with a mean relative change of \(25.94\%\). 
These results show that heterogeneous information changes not only the amount of structural information, but also the class-relevant organization of local neighborhoods.
}

\begin{table}[htbp]
\centering
{
\resizebox{\textwidth}{!}{
\begin{tabular}{lccc}
\hline
Candidate mediator retained after Step~1
& Correlation with \(\Delta Y\) (\(r\))
& \(p\)-value
& Consistent alignment (\%) \\ 
\hline

Graph density \(\rho\)
& 0.18
& 0.421
& 55.0 \\

Average degree \(K\)
& 0.24
& 0.295
& 58.5 \\

Reachability \(R\)
& 0.15
& 0.513
& 52.4 \\

Neighborhood diversity \(S\)
& 0.31
& 0.170
& 60.2 \\

\textbf{Homophily \(H\)}
& \textbf{0.78}
& \textbf{\(<0.001\)}
& \textbf{90.5} \\

\textbf{Local--global discrepancy \(D\)}
& \textbf{0.72}
& \textbf{0.002}
& \textbf{85.7} \\ 
\hline
\end{tabular}
}
}
\caption{{Second-step mediator screening based on the mediator-to-outcome relationship \(M \rightarrow Y\). 
For each candidate retained after the first step, we report its correlation with the performance improvement \(\Delta Y\), the corresponding \(p\)-value, and the percentage of cases where the direction of structural change is consistently aligned with the direction of performance improvement.}}
\label{tab:mediator_screening_step2}
\end{table}

{
Table~\ref{tab:mediator_screening_step2} reports the second-step screening results. 
Although graph density, average degree, reachability, and neighborhood diversity are affected by heterogeneous information, their changes are only weakly or unstably associated with node classification performance. 
Their correlations with \(\Delta Y\) are low, ranging from 0.15 to 0.31, and their \(p\)-values are not statistically significant. 
Their consistent alignment ratios are also close to random or only moderately above random, ranging from \(52.4\%\) to \(60.2\%\). 
These results indicate that the performance improvement from \(G_{\mathrm{Homo}}\) to \(G_{\mathrm{Homo+Hete}}\) cannot be mainly explained by simply increasing density, degree, reachability, or neighborhood diversity.
}

{
By contrast, homophily and local--global discrepancy show strong and statistically significant associations with node classification improvement. 
Homophily obtains a correlation of \(r=0.78\) with \(p<0.001\), and its structural change is consistently aligned with the performance improvement in \(90.5\%\) of cases. 
Local--global discrepancy obtains a correlation of \(r=0.72\) with \(p=0.002\), with a consistent alignment ratio of \(85.7\%\). 
These results indicate that the two factors not only respond clearly to heterogeneous information, but also explain the resulting improvement in node classification performance.
}

{
Overall, the two-step screening procedure removes factors that are either not affected by heterogeneous information or not consistently related to performance improvement. 
The remaining mediators are homophily and local--global discrepancy. 
Homophily captures whether heterogeneous information makes local neighborhoods more label-consistent, while local--global discrepancy captures whether heterogeneous information makes local neighborhoods more distinguishable from the global label distribution. 
Both factors directly reflect class separability and are therefore retained for the subsequent causal mediation decomposition.
}

\subsubsection{{Causal mediation decomposition}}

{After the two-step screening, homophily \(H\) and local--global discrepancy \(D\) are retained as the primary mediators. 
We then perform causal mediation decomposition to quantify how much of the performance improvement caused by heterogeneous information is transmitted through these selected structural factors. 
}

\begin{table}[htbp]
\centering
\caption{{Causal mediation decomposition of heterogeneous information.}}
\label{tab:mediation_decomposition}
{
\resizebox{\textwidth}{!}{
\begin{tabular}{lcccccc}
\toprule
\textbf{Mediator} 
& \textbf{ACME} 
& \textbf{ADE} 
& \textbf{Total Effect} 
& \textbf{Prop. Mediated} 
& \textbf{Bootstrap 95\% CI} 
& \textbf{Decision} \\ 
\midrule

Homophily \(H\) 
& \(0.028^{***}\) 
& \(0.030^{***}\) 
& \(0.058^{***}\) 
& \(48.3\%\) 
& \([0.021, 0.035]\) 
& Mediated \\

Local--global discrepancy \(D\) 
& \(0.019^{***}\) 
& \(0.039^{***}\) 
& \(0.058^{***}\) 
& \(32.8\%\) 
& \([0.013, 0.026]\) 
& Mediated \\

\(H + D\) jointly 
& \(0.041^{***}\) 
& \(0.017^{**}\) 
& \(0.058^{***}\) 
& \(70.7\%\) 
& \([0.034, 0.049]\) 
& Primary pathway \\

\bottomrule
\end{tabular}
}
}

\vspace{2mm}
{
\footnotesize
\textit{Note:} \(^{**}p<0.01\), \(^{***}p<0.001\).
ACME and ADE denote the average causal mediation effect and average direct effect, respectively. 
The bootstrap confidence intervals are estimated using 1,000 bootstrap resamples.
}
\end{table}

{
Table~\ref{tab:mediation_decomposition} reports the mediation decomposition results. 
Both homophily and local--global discrepancy show positive and significant mediation effects. 
For homophily, the ACME is \(0.028\), accounting for \(48.3\%\) of the total performance improvement. 
This indicates that nearly half of the benefit introduced by heterogeneous information can be explained by its effect on increasing label consistency in local neighborhoods. 
For local--global discrepancy, the ACME is \(0.019\), accounting for \(32.8\%\) of the total effect. 
This suggests that heterogeneous information also improves node classification by making local neighborhood distributions more distinguishable from the global label distribution.
}

{
When homophily and local--global discrepancy are considered jointly, the ACME increases to \(0.041\), and the proportion mediated reaches \(70.7\%\). 
Meanwhile, the ADE decreases to \(0.017\), although it remains statistically significant. 
This result shows that the two selected mediators jointly explain most of the performance gain from \(G_{\mathrm{Homo}}\) to \(G_{\mathrm{Homo+Hete}}\), while a smaller remaining direct effect may be attributed to other structural or semantic information not captured by these two mediators.
}

{
Overall, the mediation decomposition confirms that heterogeneous information improves node classification mainly by reshaping graph structure into a more class-discriminative form. 
The dominant mediation pathway is not simply the addition of more edges or relation channels, but the enhancement of class-relevant structural signals. 
Specifically, heterogeneous information increases local label consistency through homophily and strengthens local class distinctiveness through local--global discrepancy. 
These two mechanisms explain why heterogeneous information is useful for node classification under a fixed RGCN backbone.
}

\section{{Additional Experiments}}

\begin{figure}[!t]
    \centering
    \begin{subfigure}[t]{0.32\textwidth}
        \centering
        \includegraphics[width=\linewidth]{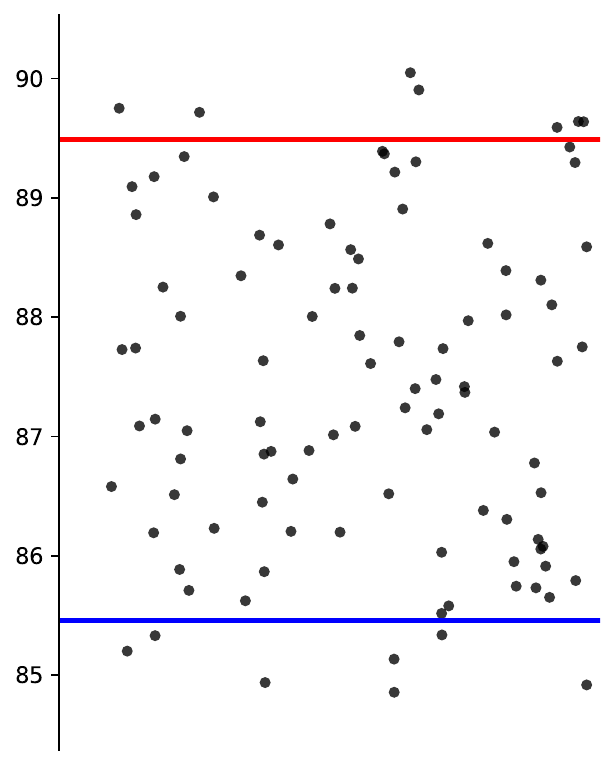}
        \caption{ACM-A}
        \label{fig:random_acma}
    \end{subfigure}
    \hfill
    \begin{subfigure}[t]{0.32\textwidth}
        \centering
        \includegraphics[width=\linewidth]{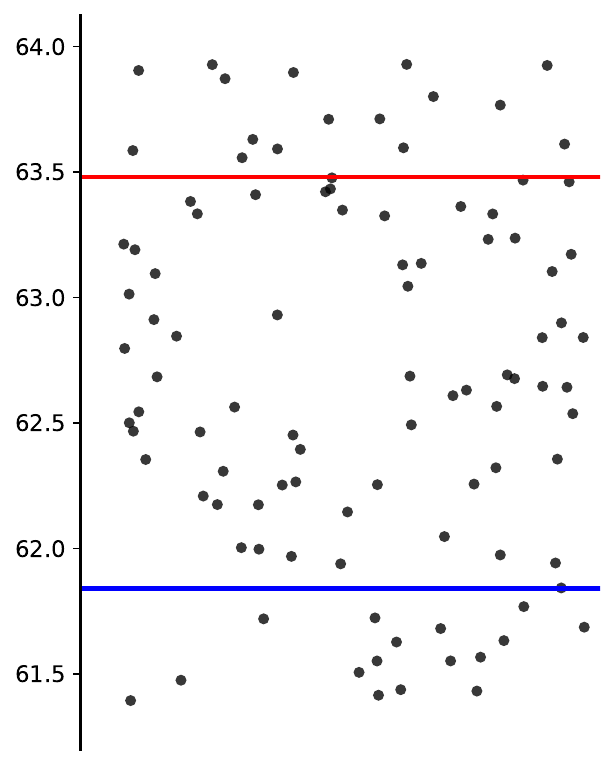}
        \caption{IMDB-A}
        \label{fig:random_imdba}
    \end{subfigure}
    \hfill
    \begin{subfigure}[t]{0.32\textwidth}
        \centering
        \includegraphics[width=\linewidth]{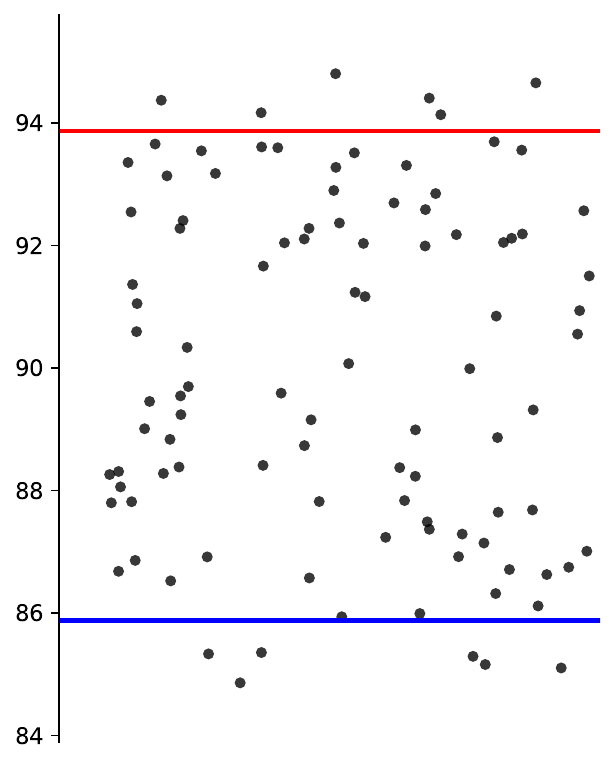}
        \caption{DBLP-A}
        \label{fig:random_dblpa}
    \end{subfigure}
    \caption{Performance comparison under randomized heterogeneous structures with a fixed RGCN architecture. For each dataset, 100 randomized heterogeneous graphs are constructed from the homogeneous projection by randomly assigning edges into the same number of relation channels as the original heterogeneous graph. The RGCN architecture and the number of relation-specific parameters remain unchanged across all settings. The black points denote randomized heterogeneous graphs, while the two horizontal lines indicate the performance of RGCN on the homogeneous projection (\textit{Homo}) and on the original heterogeneous graph (\textit{Hete}), respectively.}
    \label{fig:randomized_structure}
\end{figure}
\subsection{{Randomized Heterogeneous Structures under Fixed RGCN}}
\label{app:randomized_heterogeneous_structure}

{To further examine whether the performance gain comes merely from the increased number of relation-specific parameters, we conduct an additional randomized-structure experiment under a fixed RGCN architecture. The key idea is to keep the model architecture, parameterization, and training protocol unchanged, while only changing how edges are organized into relation channels. Specifically, for each dataset, we first collapse the original heterogeneous graph into its homogeneous projection by ignoring relation types. We then reconstruct 100 randomized heterogeneous graphs from this homogeneous projection by randomly assigning edges into the same number of relation channels as in the original heterogeneous graph. In this way, the randomized graphs have the same number of relation channels and the same RGCN parameterization as the original heterogeneous setting, but the original semantic and structural organization of relations is removed.}

{We train the same RGCN model on each randomized heterogeneous graph using the same experimental protocol. Figure~\ref{fig:randomized_structure} compares the resulting performance distributions with two reference settings: \textit{Homo}, where RGCN is applied to the homogeneous projection, and \textit{Hete}, where RGCN is applied to the original heterogeneous graph. The horizontal lines denote the performance of these two reference settings, while each black point denotes one randomized heterogeneous graph. The results show that, even under the same RGCN architecture and the same number of relation-specific parameters, different relation organizations lead to clearly different performance. Some randomized heterogeneous structures perform even worse than the \textit{Homo} setting, while a few achieve performance close to or slightly higher than the \textit{Hete} setting. This indicates that relation-specific parameterization alone is not sufficient to determine the final performance. Instead, the effectiveness of RGCN is highly sensitive to the structural information induced by relation organization. Therefore, the advantage of the original heterogeneous graph cannot be explained simply by increased model capacity; it is closely related to how heterogeneous relations reshape the graph structure.}

\begin{table*}[t]
\centering
\scriptsize
\setlength{\tabcolsep}{3pt}
\renewcommand{\arraystretch}{1.08}
\resizebox{\textwidth}{!}{
\begin{tabular}{lcccccc}
\hline
\textbf{Model} 
& \multicolumn{2}{c}{\textbf{Actor (0.29)}} 
& \multicolumn{2}{c}{\textbf{FB-MIT (0.49)}} 
& \multicolumn{2}{c}{\textbf{FB-American (0.53)}} \\
\cline{2-7}
& \textbf{MacroF1} & \textbf{MicroF1} 
& \textbf{MacroF1} & \textbf{MicroF1} 
& \textbf{MacroF1} & \textbf{MicroF1} \\
\hline
HetGNN    
& $61.48_{\pm 2.56}$ & $69.01_{\pm 2.68}$ 
& $63.14_{\pm 0.43}$ & $69.15_{\pm 0.80}$ 
& $60.47_{\pm 1.38}$ & $67.08_{\pm 0.83}$ \\

HGT       
& $63.72_{\pm 2.18}$ & $69.27_{\pm 1.34}$ 
& $63.44_{\pm 0.46}$ & $61.98_{\pm 0.33}$ 
& $59.65_{\pm 0.52}$ & $62.60_{\pm 0.41}$ \\

MAGNN     
& $66.74_{\pm 2.84}$ & $73.85_{\pm 1.39}$ 
& $72.01_{\pm 2.87}$ & $73.63_{\pm 2.66}$ 
& $71.63_{\pm 3.28}$ & $72.80_{\pm 3.01}$ \\

SimpleHGN 
& $66.94_{\pm 1.37}$ & $76.06_{\pm 0.86}$ 
& $71.89_{\pm 2.45}$ & $72.70_{\pm 3.01}$ 
& $71.88_{\pm 2.71}$ & $72.93_{\pm 2.92}$ \\

SeHGNN    
& $67.51_{\pm 0.83}$ & $76.81_{\pm 0.48}$ 
& $72.07_{\pm 2.07}$ & $74.49_{\pm 1.94}$ 
& $70.39_{\pm 3.12}$ & $73.66_{\pm 2.18}$ \\

HINormer  
& $67.12_{\pm 2.18}$ & $77.17_{\pm 2.85}$ 
& $71.23_{\pm 1.53}$ & $72.08_{\pm 2.06}$ 
& $71.03_{\pm 2.64}$ & $72.42_{\pm 2.66}$ \\

AHGNN     
& $\underline{74.89}_{\pm 0.96}$ & $\underline{82.13}_{\pm 0.08}$ 
& $\underline{73.81}_{\pm 2.07}$ & $\underline{76.32}_{\pm 1.59}$ 
& $\underline{73.75}_{\pm 1.42}$ & $\underline{75.41}_{\pm 0.80}$ \\

\hline
RGCN      
& $\mathbf{75.36}_{\pm 0.88}$ & $\mathbf{82.45}_{\pm 0.12}$ 
& $\mathbf{74.12}_{\pm 1.86}$ & $\mathbf{76.74}_{\pm 1.41}$ 
& $\mathbf{74.06}_{\pm 1.35}$ & $\mathbf{75.83}_{\pm 0.72}$ \\
\hline
\end{tabular}
}
\caption{Experimental results on heterophilic graphs.}
\label{tab:heterophilic_results}
\end{table*}

\subsection{{Performance of RGCN on Heterophilic Graphs}}
\label{subsec:rgcn_heterophilic}

{To further examine the robustness of our observations under low-homophily settings, we conduct supplementary experiments on three heterophilic heterogeneous graph datasets, including Actor, FB-MIT, and FB-American, following the benchmark setting in~\cite{chen2025adaptive}. The values in parentheses denote the homophily ratios of the datasets, where a smaller value indicates stronger heterophily. As shown in Table~\ref{tab:heterophilic_results}, conventional HGNN baselines exhibit clear performance variations on these heterophilic graphs. In contrast, after careful hyperparameter tuning, RGCN still achieves the best performance across all three datasets and all evaluation metrics. These results further support our conclusion that the effectiveness of HGNNs does not simply come from architectural complexity, but from whether heterogeneous information can be effectively exploited to provide discriminative structural signals.}

\begin{table*}[t]
\centering
\setlength{\tabcolsep}{3.5pt}
\renewcommand{\arraystretch}{1.08}
\resizebox{\textwidth}{!}{
\begin{tabular}{@{}lcccccccc@{}}
\toprule
\multirow{2}{*}{Model}
& \multicolumn{2}{c}{DBLP}
& \multicolumn{2}{c}{IMDB}
& \multicolumn{2}{c}{ACM}
& \multicolumn{2}{c}{Freebase} \\
\cmidrule(lr){2-3}
\cmidrule(lr){4-5}
\cmidrule(lr){6-7}
\cmidrule(lr){8-9}
& Macro-F1 & Micro-F1
& Macro-F1 & Micro-F1
& Macro-F1 & Micro-F1
& Macro-F1 & Micro-F1 \\
\midrule
HetGNN
& 91.76$\pm$0.43 & 92.33$\pm$0.41
& 48.25$\pm$0.67 & 51.16$\pm$0.65
& 85.91$\pm$0.25 & 86.05$\pm$0.25
& -- & -- \\
RSHN
& 93.34$\pm$0.58 & 93.81$\pm$0.55
& 59.85$\pm$3.21 & 64.22$\pm$1.03
& 90.50$\pm$1.51 & 90.32$\pm$1.54
& -- & -- \\
SAN
& 91.52$\pm$0.51 & 92.10$\pm$0.48
& 56.12$\pm$1.20 & 63.50$\pm$0.80
& 90.15$\pm$0.60 & 90.12$\pm$0.62
& 22.45$\pm$1.50 & 55.10$\pm$1.25 \\
HetSANN
& 78.55$\pm$2.42 & 80.56$\pm$1.50
& 49.47$\pm$1.21 & 57.68$\pm$0.44
& 90.02$\pm$0.35 & 89.91$\pm$0.37
& -- & -- \\
GTN
& 93.52$\pm$0.55 & 93.97$\pm$0.54
& 60.47$\pm$0.98 & 65.14$\pm$0.45
& 91.31$\pm$0.70 & 91.20$\pm$0.71
& -- & -- \\
NSHE
& 92.15$\pm$0.45 & 92.65$\pm$0.42
& 58.30$\pm$1.15 & 64.10$\pm$0.90
& 90.65$\pm$0.55 & 90.58$\pm$0.56
& 25.30$\pm$1.35 & 57.25$\pm$1.10 \\
SimpleHGN
& \underline{94.01$\pm$0.24} & \underline{94.46$\pm$0.22} & \textbf{63.53$\pm$1.36} & \textbf{67.36$\pm$0.57} & \underline{93.42$\pm$0.44} & \underline{93.35$\pm$0.45} & \underline{47.72$\pm$1.48} & \underline{66.29$\pm$0.45} \\
MHGCN
& 93.10$\pm$0.38 & 93.55$\pm$0.35
& 59.45$\pm$1.50 & 65.20$\pm$0.85
& 91.25$\pm$0.48 & 91.15$\pm$0.45
& 28.50$\pm$1.80 & 58.90$\pm$1.45 \\
RpHGNN
& 93.65$\pm$0.30 & 94.10$\pm$0.28
& 61.25$\pm$1.10 & 66.50$\pm$0.70
& 92.80$\pm$0.50 & 92.70$\pm$0.48
& 36.40$\pm$2.10 & 62.15$\pm$1.05 \\
HERO
& 93.15$\pm$0.40 & 93.60$\pm$0.38
& 60.85$\pm$1.05 & 65.80$\pm$0.65
& 91.95$\pm$0.55 & 91.85$\pm$0.52
& 31.20$\pm$1.90 & 60.10$\pm$1.20 \\
HAN
& 91.67$\pm$0.49 & 92.05$\pm$0.62
& 57.74$\pm$0.96 & 64.63$\pm$0.58
& 90.89$\pm$0.43 & 90.79$\pm$0.43
& 21.31$\pm$1.68 & 54.77$\pm$1.40 \\
MAGNN
& 93.28$\pm$0.51 & 93.76$\pm$0.45
& 56.49$\pm$3.20 & 64.67$\pm$1.67
& 90.88$\pm$0.64 & 90.77$\pm$0.65
& -- & -- \\
HGSL
& 93.85$\pm$0.28 & 94.25$\pm$0.25
& 62.40$\pm$1.25 & 66.90$\pm$0.60
& 93.10$\pm$0.45 & 93.00$\pm$0.42
& 41.50$\pm$1.65 & 64.20$\pm$0.90 \\
PHGT
& 93.45$\pm$0.35 & 93.90$\pm$0.32
& 61.50$\pm$1.15 & 66.15$\pm$0.75
& 92.20$\pm$0.52 & 92.10$\pm$0.50
& 38.60$\pm$2.00 & 63.45$\pm$1.15 \\
HOAE
& 92.85$\pm$0.42 & 93.30$\pm$0.40
& 59.10$\pm$1.35 & 64.85$\pm$0.82
& 91.45$\pm$0.60 & 91.35$\pm$0.58
& 27.80$\pm$1.55 & 58.40$\pm$1.30 \\
DMGI
& 90.55$\pm$0.65 & 91.10$\pm$0.60
& 54.20$\pm$1.80 & 61.35$\pm$1.10
& 89.65$\pm$0.85 & 89.50$\pm$0.88
& 19.50$\pm$2.10 & 52.30$\pm$1.60 \\
HINormer
& 93.75$\pm$0.25 & 94.15$\pm$0.24
& 62.85$\pm$1.05 & 67.10$\pm$0.55
& 93.25$\pm$0.40 & 93.15$\pm$0.41
& 43.20$\pm$1.45 & 65.10$\pm$0.85 \\
HGT
& 93.01$\pm$0.23 & 93.49$\pm$0.25
& 63.00$\pm$1.19 & 67.20$\pm$0.57
& 91.12$\pm$0.76 & 91.00$\pm$0.76
& 29.28$\pm$2.52 & 60.51$\pm$1.16 \\
SlotGAT
& 93.35$\pm$0.32 & 93.85$\pm$0.30
& 60.15$\pm$1.20 & 65.45$\pm$0.70
& 91.65$\pm$0.55 & 91.55$\pm$0.56
& 33.15$\pm$1.75 & 61.25$\pm$1.10 \\
HDHGR
& 92.45$\pm$0.48 & 92.95$\pm$0.45
& 58.75$\pm$1.30 & 64.30$\pm$0.85
& 90.95$\pm$0.62 & 90.85$\pm$0.60
& 26.10$\pm$1.85 & 57.85$\pm$1.35 \\
MECCH
& 93.55$\pm$0.30 & 94.05$\pm$0.28
& 61.65$\pm$1.10 & 66.25$\pm$0.65
& 92.45$\pm$0.48 & 92.35$\pm$0.45
& 35.80$\pm$1.60 & 62.55$\pm$0.95 \\
HeCo
& 93.70$\pm$0.28 & 94.12$\pm$0.26
& 62.10$\pm$1.25 & 66.75$\pm$0.60
& 92.65$\pm$0.45 & 92.55$\pm$0.46
& 37.50$\pm$1.90 & 63.80$\pm$1.05 \\
ComCH
& 93.20$\pm$0.36 & 93.70$\pm$0.34
& 59.80$\pm$1.40 & 65.10$\pm$0.78
& 91.80$\pm$0.50 & 91.70$\pm$0.52
& 32.45$\pm$1.80 & 60.85$\pm$1.25 \\
SeHGNN
& 93.95$\pm$0.22 & 94.38$\pm$0.20
& {63.40$\pm$0.95} & {67.30$\pm$0.52}
& 93.38$\pm$0.38 & 93.30$\pm$0.39
& {46.85$\pm$1.35} & {66.15$\pm$0.75} \\
\midrule
RGCN
& \textbf{94.65$\pm$0.31} & \textbf{95.00$\pm$0.28} 
& \underline{63.45$\pm$1.15} & \underline{67.32$\pm$0.50} 
& \textbf{94.11$\pm$0.30} & \textbf{94.02$\pm$0.26} 
& \textbf{48.20$\pm$1.35} & \textbf{66.75$\pm$0.40} \\
\bottomrule
\end{tabular}
}
\caption{Fair comparison of HGNN baselines on the HGB node classification benchmark under unified data splits, feature preprocessing, evaluation metrics, and tuning protocol. Results are reported as mean$\pm$std over five runs.}
\label{tab:hgb_fair_comparison}
\end{table*}

\subsection{{Fair Comparison on the HGB Benchmark}}

{To further address the concern that different baselines may have been evaluated on different datasets in prior studies, we additionally conduct a unified comparison on the Heterogeneous Graph Benchmark (HGB) \cite{lv2021we}. HGB provides standardized node classification datasets, data splits, feature preprocessing, and evaluation protocols, making it suitable for a fair comparison of heterogeneous graph neural networks under consistent experimental settings. Specifically, we evaluate all considered HGNN baselines on four HGB node classification datasets, including DBLP, IMDB, ACM, and Freebase. All models are evaluated using Macro-F1 and Micro-F1, and the results are reported as the mean and standard deviation over five independent runs. Entries marked with ``--'' indicate that the corresponding model cannot be directly evaluated on that dataset due to implementation or scalability limitations.}

{The results are reported in Table~\ref{tab:hgb_fair_comparison}. Overall, the comparison leads to several observations. First, RGCN achieves the best performance on three out of four datasets, including DBLP, ACM, and Freebase. On DBLP, RGCN obtains 94.65 Macro-F1 and 95.00 Micro-F1, outperforming the second-best SimpleHGN by 0.64 and 0.54 points, respectively. On ACM, RGCN achieves 94.11 Macro-F1 and 94.02 Micro-F1, while SimpleHGN ranks second with 93.42 Macro-F1 and 93.35 Micro-F1. On Freebase, RGCN also obtains the best results, with 48.20 Macro-F1 and 66.75 Micro-F1, again surpassing SimpleHGN, which achieves 47.72 Macro-F1 and 66.29 Micro-F1. Second, on IMDB, SimpleHGN achieves the best performance, with 63.53 Macro-F1 and 67.36 Micro-F1, while RGCN ranks second with 63.45 Macro-F1 and 67.32 Micro-F1. The performance gap between the two models on IMDB is marginal, suggesting that RGCN remains highly competitive even when it is not the top-performing model.}

{These results strengthen our main conclusion from the reproduction study. Under a unified benchmark and consistent evaluation protocol, more sophisticated HGNN architectures do not consistently outperform the simple RGCN model. Although several recent models, such as SeHGNN, HINormer, HGSL, and HGT, achieve competitive results on specific datasets, their improvements are not consistent across all HGB datasets. In contrast, RGCN maintains strong and stable performance across different graph domains. This provides additional evidence that the performance differences among HGNNs cannot be solely attributed to architectural sophistication, further motivating our subsequent analysis of heterogeneous information itself.}

\section{Conclusion}
To address the lack of rigorous evidence on the effectiveness of HGNNs for node classification, we conduct a comprehensive causal analysis that separates model architecture from heterogeneous information.
The results show that model architecture and complexity have no causal effect on node classification performance.
In contrast, heterogeneous information exerts a positive causal effect by reshaping graph structure, increasing homophily, and enlarging the difference between local and global label distributions, which makes node classes more distinguishable and improves predictive accuracy.



\bibliographystyle{elsarticle-num} 
\bibliography{main}
\end{document}